\documentclass[10pt,twocolumn,letterpaper]{article}

\usepackage[pagenumbers]{cvpr} 

\usepackage{microtype}
\usepackage{marvosym}
\usepackage{multirow}
\usepackage{multicol}
\usepackage{tabularx}
\usepackage{array}
\usepackage{gensymb}
\usepackage[table]{xcolor}

\definecolor{cvprblue}{rgb}{0.21,0.49,0.74}
\usepackage[pagebackref,breaklinks,colorlinks,allcolors=cvprblue]{hyperref}

\def\paperID{7614} 
\def\confName{CVPR}
\def\confYear{2026}

\title{Reliev3R: Relieving Feed-forward 3D Reconstruction 

from Multi-View Geometric Annotations}

\author{
Youyu Chen\textsuperscript{1\#}~~~~~Junjun Jiang\textsuperscript{1\Letter}~~~~~Yueru Luo\textsuperscript{3\#}~~~~~Kui Jiang\textsuperscript{1}\\
Xianming Liu\textsuperscript{1}~~~~~Xu Yan\textsuperscript{2}~~~~~Dave Zhenyu Chen\textsuperscript{2}\\
\vspace{-3pt} \small \textsuperscript{1}Harbin Institute of Technology 
\quad \textsuperscript{2}Huawei \quad \textsuperscript{3}The Chinese University of Hong Kong, Shenzhen \\
\small \textsuperscript{\#}{Work done during internship at Huawei} \quad \textsuperscript{\Letter}{Corresponding Author}
}

\begin{document}
\newcommand\blfootnote[1]{%
  \begingroup
  \vspace{-2mm}
  \renewcommand\thefootnote{}\footnote{#1}%
  \addtocounter{footnote}{-1}%
  \endgroup
}

\begin{figure}
    \vspace{0mm}
    \captionsetup{singlelinecheck=false}
    \twocolumn[{
        \renewcommand\twocolumn[1][]{#1}
        \maketitle
        
        \centering
        \setlength\tabcolsep{0pt}
        \includegraphics[width=\linewidth]{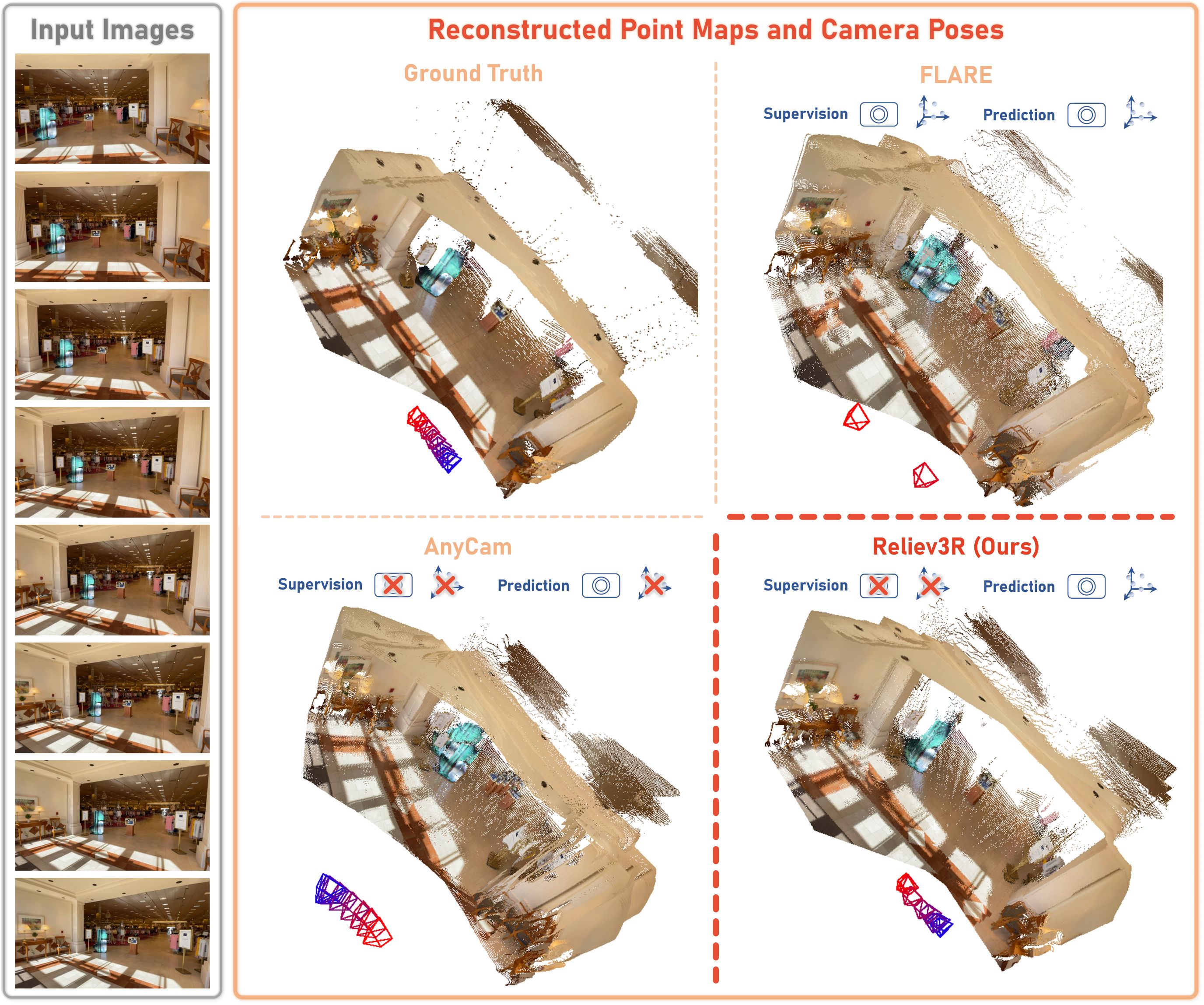}
        \caption{
            In this paper, we propose Reliev3R, the first learning paradigm to train a \textbf{F}eed-\textbf{f}orward \textbf{R}econstruction \textbf{M}odel (FFRM) \textbf{\textit{from scratch}} without reliance on multi-view geometric annotations. 
            To learn multi-view geometric knowledge, Reliev3R engages pseudo monocular relative depths and pseudo image sparse correspondences with multi-view geometry constraints as a weakly-supervised learning objective. 
            As shown above in the figure, Reliev3R surpasses early FFRMs (\eg, FLARE~\cite{zhang2025flare}) and weakly-supervised camera pose estimation models (\eg, AnyCam~\cite{wimbauer2025anycam}) in the overall performance. 
        }
        \vspace{4mm}
        \label{fig:teaser}
    }]
\end{figure}

\clearpage
\begin{abstract}


With recent advances, \textbf{F}eed-\textbf{f}orward \textbf{R}econstruction \textbf{M}odels (FFRMs) have demonstrated great potential in reconstruction quality and adaptiveness to multiple downstream tasks. However, the excessive reliance on multi-view geometric annotations, e.g. 3D point maps and camera poses, makes the fully-supervised training scheme of FFRMs difficult to scale up.
In this paper, we propose \textbf{Reliev3R}, a weakly-supervised paradigm for training FFRMs from scratch without cost-prohibitive multi-view geometric annotations. 
Relieving the reliance on geometric sensory data and compute-exhaustive structure-from-motion preprocessing, our method draws 3D knowledge directly from monocular relative depths and image sparse correspondences given by zero-shot predictions of pretrained models.
At the core of Reliev3R, we design an ambiguity-aware relative depth loss and a trigonometry-based reprojection loss to facilitate supervision for multi-view geometric consistency.
Training from scratch with the less data, Reliev3R catches up with its fully-supervised sibling models, taking a step towards low-cost 3D reconstruction supervisions and scalable FFRMs.

\end{abstract}    
\section{Introduction}
\label{sec:intro}

3D reconstruction, of which the ultimate goal is to reconstruct the 3D content of a scene directly from a set of images, has been investigated by the community for decades. 
Recent advances in \textbf{F}eed-\textbf{f}orward \textbf{R}econstruction \textbf{M}odels (FFRMs) demonstrate the feasibility of an end-to-end paradigm that maps 2D images to 3D contents such as point maps and camera poses~\cite{wang2024dust3r, wang2025vggt, wang2025pi, keetha2025mapanything}. 
Beyond reducing the time cost of 3D reconstruction from minutes to merely seconds, FFRMs also empower down-stream 3D applications such as text-3D grounding~\cite{ma2025spatialllm} and embodied AI~\cite{ge2025vggt} with its 3D-knowledge-rich image encoder.

Despite their impressive achievements, current FFRMs heavily rely on large-scale multi-view geometric annotations, including camera parameters, dense point maps, \etc. Such annotations must satisfy strict 3D consistency: point maps across views must align into a coherent 3D structure. 
In practice, these labels are obtained through a multi-stage pipeline built upon Structure-from-Motion (SfM)~\cite{schonberger2016structure, pan2024global} followed by multi-view stereo (MVS)~\cite{fischler1981random, triggs1999bundle, izquierdo2025mvsanywhere, wang2025moge}. 
The pipeline is computationally expensive, often brittle in low-texture or challenging scenes, and difficult to scale. 
As a consequence, the performance of FFRMs is limited not by model design but by the scarcity and cost of producing large-scale, high-consistency geometric supervision.



A key observation is that these multi-view geometric annotations are not the essence of reconstruction. The raw multi-view inputs already contain all the necessary cues for geometry: depth–appearance relationships, multi-view correspondences, and pose-induced reprojection structure. Supervising FFRMs with SfM/MVS annotations can therefore be interpreted as embedding the traditional reconstruction pipeline inside a transformer through ground-truth labels. This naturally raises an essential question: instead of supervising FFRMs with outputs from SfM/MVS, can we directly learn the intrinsic geometric principles from the multi-view inputs themselves, without relying on heavy geometric annotations?


With this insight, we propose Reliev3R, the first weakly supervised training paradigm that enables learning FFRMs from scratch without multi-view geometric annotations~\cite{bochkovskii2024depth, karaev2024cotracker}. Reliev3R reconstructs a 3D-aligned depth map per view and introduces two key forms of lightweight supervision. First, monocular relative depth maps predicted by pretrained priors are used as pseudo-labels to constrain the shape of each depth distribution. To address the inherent multi-view inconsistencies in monocular estimation, we develop an ambiguity-aware scale-invariant depth loss that automatically downweights unreliable regions such as sky or reflective surfaces. Second, sparse 2D correspondences from an off-the-shelf matcher provide geometric anchors across views, enabling Reliev3R to register depth predictions into a globally aligned 3D space. The predicted camera poses and aligned depths are jointly optimized using a differentiable trigonometry-based reprojection loss, while camera intrinsics are assumed known. This formulation eliminates the dependency on SfM/MVS annotations and allows scaling to larger, more diverse training data.

We conduct experiments to compare Reliev3R and other FFRMs that are fully-supervised with multi-view geometric annotations. 
Reliev3R matches against and even outperforms early FFRMs such as MVDUSt3R~\cite{tang2025mv} and FLARE~\cite{zhang2025flare}, while delivering comparable reconstruction accuracies with SOTA FFRMs~\cite{ge2025vggt, wang2025pi, keetha2025mapanything}.
Notably, in terms of camera pose estimation, Reliev3R significantly outperforms AnyCam~\cite{wimbauer2025anycam}, which is expertise at camera parameter estimation and also not supervised with multi-view geometric annotations. 
All results above showcase the superiority of Reliev3R and the feasibility of learning multi-view geometry without a heavy SfM-based reconstruction pipeline. 
To summarize our contributions, 

\begin{itemize}
    \item We propose Reliev3R, the first weakly-supervised learning paradigm that trains a feed-forward reconstruction model \textit{from scratch} without multi-view geometric annotations. 
    \item We introduce an ambiguity-aware scale-invariant depth loss and a trigonometry-based reprojection loss, which cooperate to register view-wise depth and camera pose predictions in 3D world coordinate. 
    \item Through extensive experiments, we show that Reliev3R approaches or surpasses the performance of FFRMs trained with multi-view geometric annotations, while requiring far less curated geometric data.
\end{itemize}

\begin{figure*}[t]
    \centering
    \captionsetup{singlelinecheck=false}
    \includegraphics[width=\linewidth]{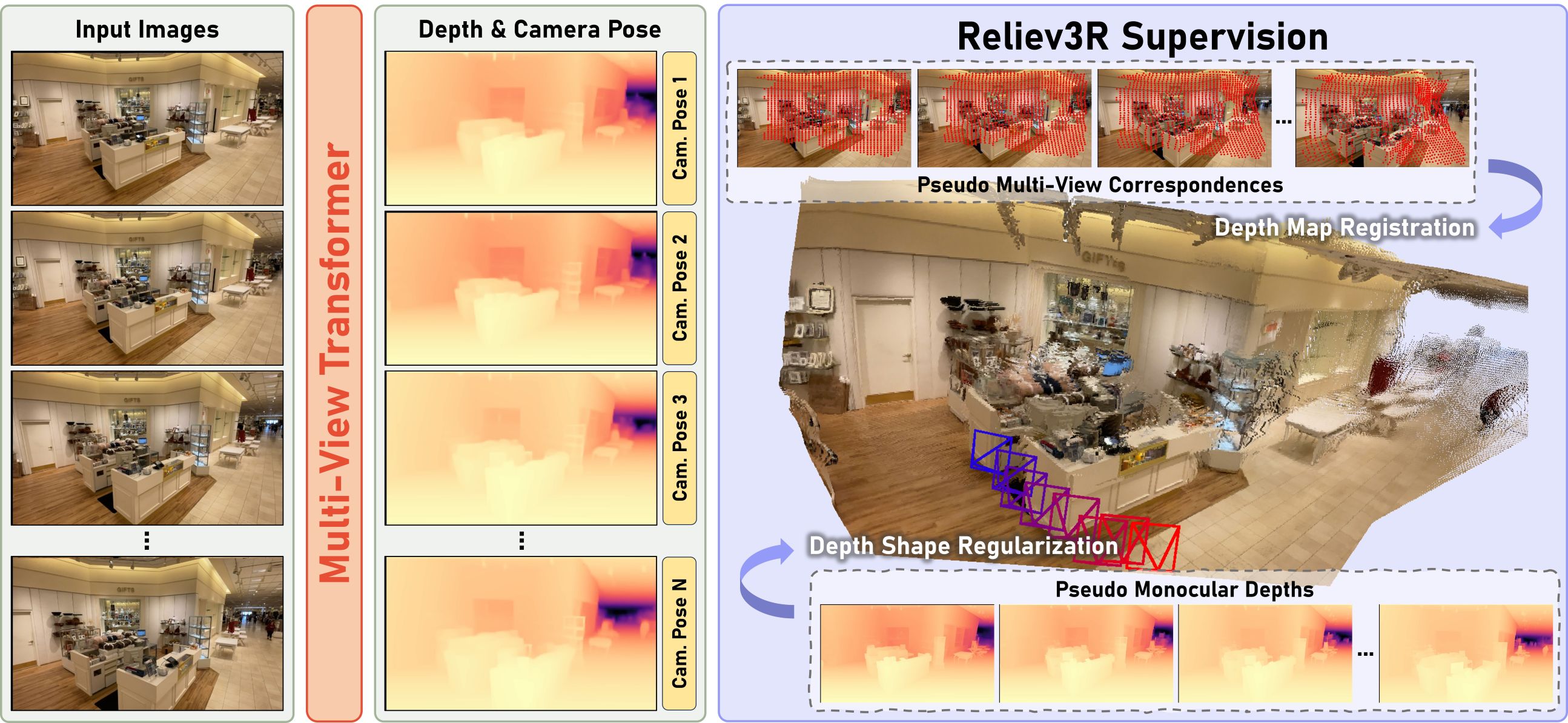}
    \caption{
        Similar to prior FFRMs, Reliev3R performs geometry reconstruction in a single forward pass given a group of images. 
        However, the supervision of Reliev3R doesn't depend on any multl-view geometric annotations (\eg, ground truth of point maps and camera poses which are registered in 3D world coordinate). 
        Specifically, instead of directly predicting the point maps registered in world coordinate, Reliev3R predicts view-wise depth maps, which are regularized with pseudo relative depth. 
        To learn the registration of depth maps and camera poses in world coordinate, Reliev3R draws supervision signal from pseudo image correspondences. 
        Pseudo annotations used by Reliev3R are produced with pretrained expert models (see \cref{sec:experiments:details}). 
        Experiments in \cref{sec:experiments} demonstrate that Reliev3R catches up with and even surpasses some of FFRMs supervised with multi-view geometric annotations. 
    }
    \label{fig:pipeline}
\end{figure*}

\section{Related Works}
\label{sec:related_works}

\vspace{-2mm}
\paragraph{Evolvement of 3D Reconstruction.}
With image correspondences derived from handcrafted feature descriptors~\cite{lowe2004distinctive, tuytelaars2008local}, conventional 3D reconstruction methods estimate camera parameters as the base of subsequent reconstruction~\cite{fischler1981random, triggs1999bundle}. 
As a milestone, structure-from-motion (SfM) methods~\cite{schonberger2016structure} integrate image matching, camera estimation and point cloud reconstruction into a sophisticated pipeline, which remains to be widely applied in up-to-date open-sourced 3D datasets to produce annotation for camera parameters~\cite{ling2024dl3dv, dai2017scannet, zhou2018stereo}. 
While SfM methods struggles to produce dense point cloud for their low-efficiency, the emergence of deep-learning encourages the community to investigate learnable multi-view stereo (MVS) methods that aim to train dense depth prediction networks inputting images and camera parameters~\cite{yao2018mvsnet, yao2019recurrent}. 
More recently, differentiable rendering~\cite{kato2020differentiable} facilitates the development of NeRF~\cite{mildenhall2021nerf, barron2022mip, muller2022instant} and 3DGS~\cite{kerbl20233d, yu2024mip}, which propose to optimize parameters describing the 3D space in a scene-wise manner. 
Despite the SOTAs of 3DGS are now capable to reconstruct a scene in minutes~\cite{mallick2024taming, fang2024mini, chen2025dashgaussian}, the time delay is not acceptable for applications requiring real-time reconstruction like automatic driving~\cite{zhang2025visionpad} and embodied AI~\cite{ge2025vggt}.
\vspace{-4mm}
\blfootnote{\textsuperscript{\Letter}Corresponding author. E-mail: \tt{jiangjunjun@hit.edu.cn}.}

\vspace{-4mm}
\paragraph{Feed-Forward Reconstruction Models.}
The emergence of DUSt3R~\cite{wang2024dust3r}, along with its subsequent sibling improvements~\cite{leroy2024grounding, tang2025mv, wang2025continuous, zhang2024monst3r}, proves end-to-end learnable feed-forward reconstruction to be feasible from image pairs to 3D contents, \eg, 3D point maps and camera poses. 
These feed-forward reconstruction models (FFRMs), starting from merely capable to infer on two views, are now expanded to infer on thousands of images in seconds~\cite{tang2025mv, yang2025fast3r, zhang2025flare, wang20243d}. 
The improvement on reconstruction precision of FFRMs is accompanied with the expansion of training data~\cite{ge2025vggt, wang2025pi, keetha2025mapanything}, which is composed of multi-view images labeled with multi-view geometric annotations to make full-supervision on the model output. 
However, the amount of available training data seems to meet its bound, that high-quality annotations for FFRM supervision is cost-prohibitive to acquire. 
This encourages us to explore a new supervision paradigm for FFRM training, which can be relieved from expensive multi-view geometric annotations and easily scale up with massive unlabeled video data.

\vspace{-4mm}
\paragraph{Other Feed-Forward 3D Models.}
Under the impetus from FFRMs, 3D tasks such as novel-view synthesis and camera parameter estimation beside point map reconstruction can now be performed in a feed-forward approach as well, most of which are finetuned on pretrained FFRM checkpoints.
For example, feed-forward Gaussian Splatting models~\cite{ye2024no, jiang2025anysplat} typically concatenate a Gaussian primitive prediction head following pretrained FFRMs, supervised with differentiable rendering given ground truth camera parameters. 
Only a minority of researches study with supervising a feed-forward 3D model from scratch without multi-view geometric annotations. 
For example, AnyCam~\cite{wimbauer2025anycam} predicts camera parameters alone inputting images, pseudo depths and pseudo optical flows, supervised with multi-view geometric constraints engaging depths, optical flows and camera poses. 
RayZer~\cite{jiang2025rayzer} is a feed-forward neural rendering model for novel-view synthesis. 
Supervised with differentiable rendering on input images only, RayZer struggles with producing meaningful camera poses and geometry reconstruction. 
As comparison, our Reliev3R simultaneously predicts registered geometry of the scene and camera poses for the input views, presenting a new paradigm for FFRM training with potential to benefit all other 3D tasks as FFRMs do. 

        

\begin{figure*}[t]
    \centering
    \captionsetup{singlelinecheck=false}
    \includegraphics[width=\linewidth]{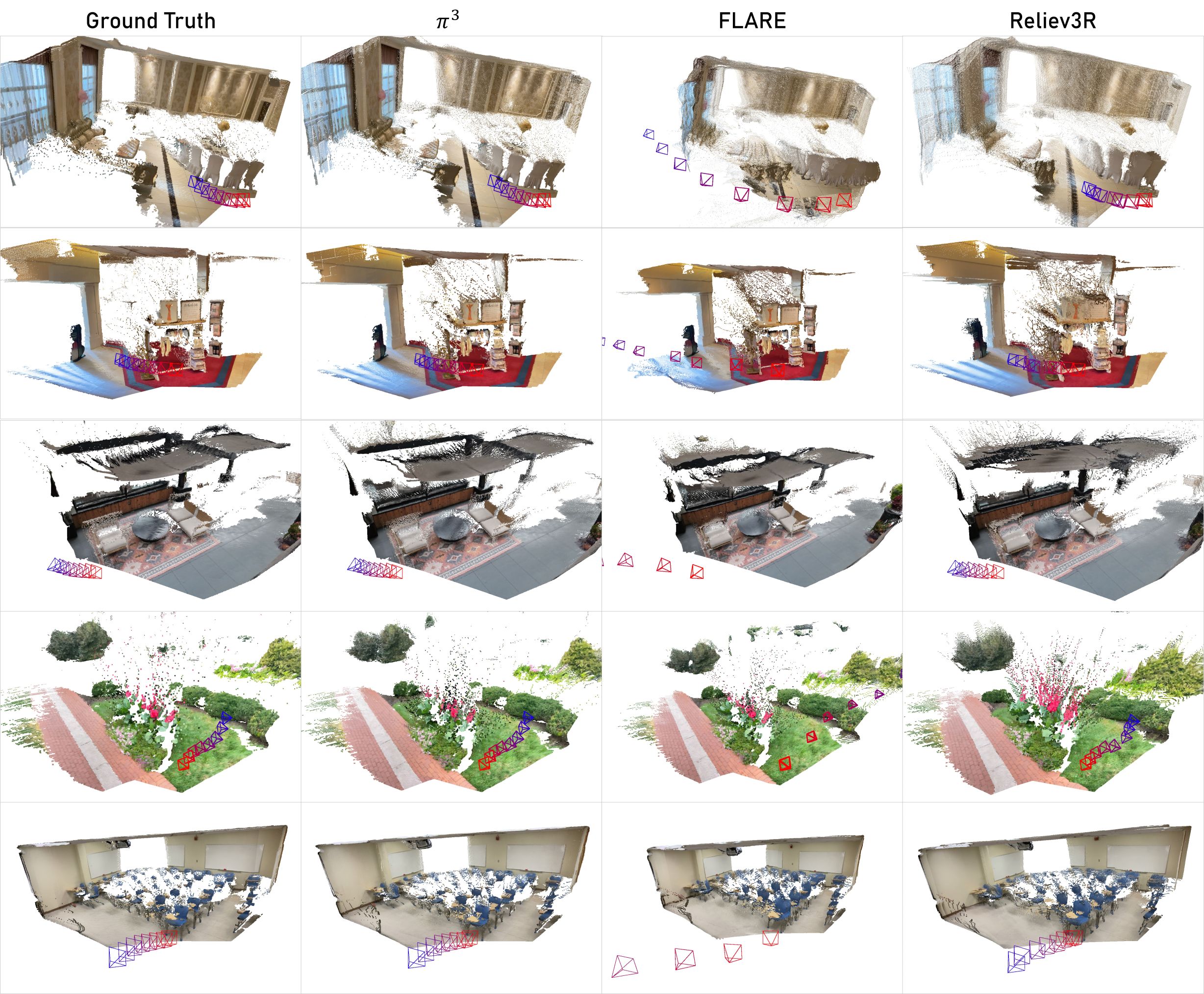}
    \caption{
        Point map and camera pose visualization on DL3DV-benchmark dataset~\cite{ling2024dl3dv}. 
        As shown above in the figure, Reliev3R delivers a visually comparable construction accuracy with the $\pi^3$~\cite{wang2025pi} while surpassing FLARE~\cite{zhang2025flare} in the overall performance. 
        The camera pose of FLARE above is adopted directly from the model prediction instead of solved with PnP~\cite{fischler1981random} based on point maps.
    }
    \label{fig:zeroshot}
\end{figure*}

\section{Method}
\label{sec:method}

In this section, we introduce our Reliev3R in detail. 
In \cref{sec:method:revisit}, we revisit the supervision of FFRMs and points out the key difference between the training paradigm of Reliev3R and other FFRMs. 
In \cref{sec:method:shape_reg} and \cref{sec:method:mvgl}, we delineate the proposed training objectives for Reliev3R, which enables the supervision for FFRMs without multi-view geometric annotations.

\subsection{Revisiting Supervisions of FFRMs}
\label{sec:method:revisit}

The success of FFRMs attributes to the high-quality multi-view geometric annotations of training data. 
By learning these annotations, FFRMs essentially compress the reconstruction progress from a complicated multi-view geometric prediction pipeline into a single inference through a deep-learning model. 
We denote this process as, 
\begin{equation}
    \label{eq:1}
    \mathcal{A}: \mathrm{I} \rightarrow (\mathrm{P}, \mathrm{C}), \ 
    \mathcal{F} = \mathcal{A}.
\end{equation}
where $\mathcal{A}$ denotes the annotation pipeline, $\mathrm{I}=\{\mathrm{I}_i\}_{i=1}^N$ denotes $N$ images photographing the scene to be annotated, $\mathrm{P}=\{\mathrm{P}_i\}_{i=1}^N$ denotes the annotated point maps of images, $\mathrm{C}=\{\mathrm{C}_i\}_{i=1}^N$ denotes the annotated camera poses, and $\mathcal{F}$ denotes the FFRMs to be trained with the annotated data. 
Such a supervised learning paradigm makes it straight-forward but cost-prohibitive to scale up like large language models and 2D foundation models, as the acquirement of annotations demands massive efforts. 

By decomposing the annotation pipeline of FFRM training data, we locate the major annotation cost at the estimation from image correspondences to dense 3D point maps and camera poses, which can be denoted as, 
\begin{equation}
    \label{eq:2}
    \mathcal{A}=\mathcal{R} \circ \mathcal{M}, \ 
    \mathcal{M}: \mathrm{I} \rightarrow \mathrm{M}, \ 
    \mathcal{R}: \mathrm{M} \rightarrow (\mathrm{P}, \mathrm{C}), 
\end{equation}
where $\mathcal{M}$ denotes the image matching process to obtain image correspondences, $\mathcal{R}$ denotes the reconstruction process to solve camera poses and dense point maps from image correspondences, and $\mathrm{M}=\{\mathrm{M_{i,j}}\}_{i,j=1}^N$ denotes the pair-wise image correspondences of $\mathrm{I}$. 

Since image correspondences are the origin of all other multi-view geometric annotations, a question is naturally raised: can image correspondences alone provide necessary supervisions for training FFRMs, eliminating the redundant preprocessing of 3D annotations?
This question mark paves the way towards the foundation of Reliev3R. 
However, the question is non-trivial since the image correspondences are typically sparse and noisy, which is not capable to produce dense point maps. 
To mitigate this problem, we propose to regularize the dense geometric prediction of FFRMs with monocular relative depth predicted by pretrained models. 
Constrained with multi-view image correspondences and multi-view geometry, the predicted depth maps of FFRMs can be aligned in 3D space to produce registered 3D point maps of the scene. 

With the above analysis, now we summarize Reliev3R mathematically as, 
\begin{equation}
    \label{eq:3}
    \mathcal{F}_\mathrm{R}: \mathrm{I} \rightarrow (\mathrm{D}, \mathrm{C}), \ 
    \mathcal{F}_\mathrm{R} = \mathcal{S} \circ (\mathcal{M} \times \mathcal{D}), \ 
    \mathcal{D}: \mathrm{I} \rightarrow \mathrm D, 
\end{equation}
where $\mathcal{F}_\mathrm{R}$ is the mapping that Reliev3R learns, $\mathrm{D} = \{ \mathrm{D}_i \}_{i=1}^N$ is the view-wise dense depth maps predicted by Reliev3R, and $\mathcal{D}$ is the pretrained monocular depth model introduced to produce relative depth information. 
$\mathcal{S}$ is the supervisions applied upon image correspondences and monocular depth maps to ensure multi-view geometric consistency and register predicted depth maps in 3D space, which is delineated in the following sections.  
Given that $\mathrm{D}$ is equivalent to $\mathrm{P}$ when camera intrinsics are known, the target of Reliev3R is to minimize the difference between $\mathcal{F}$ and $\mathcal{F}_\mathrm{R}$ as much as possible.

\subsection{Shape Regularization of Depth Maps}
\label{sec:method:shape_reg}

Ideally, Reliev3R should be supervised with multi-view geometric consistent depth maps that are naturally aligned when back-projected into world coordinate with cameras. 
However, this is infeasible without multi-view geometric annotations, which are the very objects that Reliev3R is purpose-built to sidestep. 
Despite monocular metric depth models claim to produce depth maps in metric coordinate, there is still inevitably multi-view inconsistency in both global scale and local details of monocular depth maps when multiple predictions are made for the same scene. 

To address the multi-view inconsistency in global depth scale, we adopt pretrained monocular depth models to only regularize the scale-invariant shape of depth maps (relative depth), which can be easily accomplished with canonical scale-invariant depth loss~\cite{yang2024depth}. 
Registration of depth maps in world coordinate is left to \cref{sec:method:mvgl}.
On the other hand, the inconsistency in local details should also not be overlooked. 
This is significant for the subsequent depth map registration, because depth maps with an inconsistent local shape can never be registered in world coordinate with any scale and shift. 
For example, the depth of sky is meaningless, and should be ignored when the depth maps are registered. 
To this end, we introduce an ambiguity-aware scale-invariant depth loss $\mathcal{L}_\text{d}$, which is self-adaptive to recognize the regions to be ignored for depth shape regularization. 
Given a predicted depth map $\hat{\mathrm{D}}_i$ and pseudo depth annotation $\mathrm{D}_i$ from pretrained monocular depth model of the $i$-th image view, $\mathcal{L}_\text{d}$ is denoted as, 
\begin{equation}
    \label{eq:4}
    \mathcal{L}_\text{d} = \mathrm{W}_i \cdot |\Gamma(\hat{\mathrm{D}}_i, \text{sg}(\mathrm{W}_i)) - \Gamma(\mathrm{D}_i, \text{sg}(\mathrm{W}_i))| - \alpha\log(\mathrm{W}_i), 
\end{equation}
where $\mathrm{W}_i$ denotes the estimated confidence map from Reliev3R, $\Gamma$ denotes the \textbf{w}eighted \textbf{m}edian \textbf{a}bsolute \textbf{d}eviation (WMAD), and $\text{sg}$ denotes the gradient detach operation. 
$\mathrm{W}_i$ is restricted within $(0, 2)$, that $\mathrm{W}_i\rightarrow 2$ encourages the model to explore complicated regions while $\mathrm{W}_i\rightarrow 0$ implicates multi-view inconsistent depth to be ignored. 
We adopt WMAD to shield the influence from low-confidence pixels to the relative depth normalization process. 
Please move to the supplementary for further details about $\mathcal{L}_\text{d}$.

\subsection{Multi-View Geometry Registration Learning}
\label{sec:method:mvgl}

With the shape of depth maps regularized, the rest is to estimate the camera pose of each view and register the depth maps in the world coordinate. 
We denote the image correspondences of a view pair $(i,j)$ as $\mathrm{M}_{i,j}=\{ (\mathrm{u}_{i,k}, \mathrm{u}_{j,k}) \}_{k=1}^K$, where $\mathrm{u}_{i,k}$ denotes the $k$-th corresponded pixel of the pair in view $i$ and $K$ denotes the number of total corresponded pixels.
It is natural that $\mathrm{p}_{i,k}$ should be equal with $\mathrm{p}_{j,k}$, with 
\begin{equation}
    \label{eq:5}
    \mathrm{p}_{i,k}=\mathrm{C}_i \times \pi(\mathrm{u}_{i,k}, \mathrm{d}_{i,k}), 
\end{equation}
where $\mathrm{d}_{i,k}$ denotes predicted depth of $\mathrm{u}_{i,k}$, $\pi$ denotes the back-projection from pixels in image space to 3D points in camera coordinate via known intrinsics, and the 3D point in world coordinate $\mathrm{p}_{i,k}$ of pixel $\mathrm{u}_{i,k}$ is obtained via the camera-to-world transformation $\mathrm{C}_i$ of view $i$. 
Minimizing distance between $\mathrm{p}_{i,k}$ and $\mathrm{p}_{j,k}$ produces gradient to optimize $\mathrm{C}_i$ and register $\mathrm{d}_{i,k}$ with $\mathrm{d}_{j,k}$. 

However, we find the optimization crashes with a trivial minimization on $\sum_{i,j,k}{||\mathrm{p}_{i,k} - \mathrm{p}_{j,k}||_2}$. 
This is because points far from the camera, which have big numeric in coordinate, tend to cause a large error in this formulation. 
It tempts $\mathrm{p}$ to fall towards camera center, resulting in distortion of the registered point cloud. 
And a trivial reprojection error in pixel space dosen't work as well, because it cannot produce right gradient to optimize the camera pose when $\mathrm{p}_{j,k}$ falls behind the image plane of view $i$. 

As the solution, we propose a trigonometry-based reprojection loss, which is essentially a reprojection error but optimized in 3D space. 
Denoting the camera center of view $i$ as $\mathrm{t}_i$, we optimize the following objective, 
\begin{equation}
    \label{eq:6}
    \mathcal{L}_\text{rgst} = \sum_{i,j,k} \langle \mathrm{p}_{i,k} - \mathrm{t}_i,\ \mathrm{p}_{j,k} - \mathrm{t}_i \rangle, 
\end{equation}
where the operator $\langle \cdot, \cdot \rangle$ is to calculate the angle between two vectors. 
This learning objective concentrates on minimizing the difference in direction of vectors. 
The direction of $\mathrm{p}_{i,k} - \mathrm{t}_i$ is only relevant with the rotation of view $i$ camera since its intrinsic is known and fixed. 
And the direction of $\mathrm{p}_{j,k} - \mathrm{t}_i$ is determined by the relative position between $\mathrm{p}_{j,k}$ and view $i$ camera center $\mathrm{t}_i$, which produces the gradient to change the scale of $\mathrm{D}_j$ for depth map registration. 
Please refer to the supplementary for more details about $\mathcal{L}_\text{rgst}$.

Finally, the training objective of Reliev3R can be summarized as, 
\begin{equation}
    \label{eq:7}
    \mathcal{L}=\mathcal{L}_\text{d}+\lambda\cdot \mathcal{L}_\text{rgst}, 
\end{equation}
where $\lambda$ is a weighting factor set to $0.5$ in this paper by default. 

\begin{table*}[t!]
    \centering
    \setlength\tabcolsep{8pt}
    \captionsetup{singlelinecheck=false}
    \caption{
        Evaluation on DL3DV-benchmark dataset~\cite{ling2024dl3dv} for 8-view reconstruction. 
        \textbf{Reliev3R outperforms early FFRMs trained with MVS annotations} like MVDUSt3R~\cite{tang2025mv} and FLARE~\cite{zhang2025flare}, demonstrating the advantage and potential of our method.
        We report the absolute relative error (rel) and inlier ratio at a relative threshold of 10\% ($\tau$) to evaluate the reconstructed point maps and depth maps. 
        We also report the average aligned trajectory error (ATE) and area under curve at an error threshold of 30$\degree{}$ to evaluate camera pose estimation. 
        $\pi^{3\dag}$ denotes the result of training $\pi^{3}$~\cite{wang2025pi} from scratch on DL3DV-10K~\cite{ling2024dl3dv} dataset. 
        We display if based methods are trained with/without multi-view geometric (MVG) annotations, and classify them by whether DL3DV-10K dataset is used for training into `Yes', `Alone' and `No'. 
        SOTAs trained with 20 times of data for Reliev3R are colored as \textcolor{lightgray}{gray}. 
        \colorbox{red!50}{Best}, \colorbox{orange!50}{second best} and \colorbox{yellow!25}{third best} are colored respectively. 
    }
    \begin{tabular}{ c c c cc cc cc } 
        \toprule
        {\multirow{2}{3cm}{\centering Method}} & {\multirow{2}{2cm}{\centering Training with DL3DV-10K }} & {\multirow{2}{2cm}{\centering MVG Annotations}} & 
        \multicolumn{2}{c}{Point Map} & \multicolumn{2}{c}{Camera Pose} & \multicolumn{2}{c}{Depth Map} \\
                              & & & rel $\downarrow$ & $\tau\uparrow$ &  ATE $\downarrow$ & AUC $\uparrow$ & rel $\downarrow$ & $\tau\uparrow$ \\

        \midrule

        \multicolumn{1}{l}{\textcolor{lightgray}{Map-Anything~\cite{keetha2025mapanything}}} & \textcolor{lightgray}{Yes}            &  \textcolor{lightgray}{w}  &          \textcolor{lightgray}{0.051} &        \textcolor{lightgray}{0.911} &           \textcolor{lightgray}{0.004} &        \textcolor{lightgray}{87.088} &          \textcolor{lightgray}{0.045} &        \textcolor{lightgray}{0.896} \\
        \multicolumn{1}{l}{\textcolor{lightgray}{$\pi^3$~\cite{wang2025pi}}}       & \textcolor{lightgray}{No}             &  \textcolor{lightgray}{w}  &          \textcolor{lightgray}{0.056} &        \textcolor{lightgray}{0.927} &           \textcolor{lightgray}{0.003} &        \textcolor{lightgray}{92.746} &          \textcolor{lightgray}{0.054} &        \textcolor{lightgray}{0.916} \\
        \multicolumn{1}{l}{\textcolor{lightgray}{VGGT~\cite{wang2025vggt}}}        & \textcolor{lightgray}{Yes}            &  \textcolor{lightgray}{w}  &          \textcolor{lightgray}{0.061} &        \textcolor{lightgray}{0.905} &           \textcolor{lightgray}{0.003} &        \textcolor{lightgray}{94.795} &          \textcolor{lightgray}{0.059} &        \textcolor{lightgray}{0.899} \\
        \multicolumn{1}{l}{\textcolor{lightgray}{CUT3R~\cite{wang2025continuous}}} & \textcolor{lightgray}{Yes}            &  \textcolor{lightgray}{w}  &          \textcolor{lightgray}{0.073} &        \textcolor{lightgray}{0.815} &           \textcolor{lightgray}{0.005} &        \textcolor{lightgray}{89.432} &          \textcolor{lightgray}{0.072} &        \textcolor{lightgray}{0.801} \\
        \multicolumn{1}{l}{$\pi^{3\dag}$}                    & Alone       &  w  &          \cellcolor{red!50}0.057 &        \cellcolor{red!50}0.897 &           \cellcolor{red!50}0.013 &        \cellcolor{orange!50}63.314 &          \cellcolor{red!50}0.047 &        \cellcolor{red!50}0.890 \\
        \multicolumn{1}{l}{MVDUSt3R~\cite{tang2025mv}}      & No             &  w  &          0.593 &        0.233 &           0.458 &         2.529 &          0.690 &        0.134 \\
        \multicolumn{1}{l}{FLARE~\cite{zhang2025flare}}     & Yes            &  w  &          \cellcolor{yellow!25}0.134 &        \cellcolor{orange!50}0.771 &           0.338 &        \cellcolor{red!50}80.905 &          0.376 &        0.302 \\

        \hline
        
        \multicolumn{1}{l}{AnyCam~\cite{wimbauer2025anycam}} & No           & w/o &          0.262 &        0.490 &           \cellcolor{yellow!25}0.023 &        29.527 &          \cellcolor{yellow!25}0.181 &        \cellcolor{yellow!25}0.400 \\
        \multicolumn{1}{l}{Reliev3R(ours)}                   & Alone       & w/o &          \cellcolor{orange!50}0.122 &        \cellcolor{yellow!25}0.663 &           \cellcolor{orange!50}0.018 &        \cellcolor{yellow!25}49.426 &          \cellcolor{orange!50}0.115 &        \cellcolor{orange!50}0.657 \\
        
        \bottomrule
    \end{tabular}
    \label{tab:main-result}
\end{table*}

\section{Experiments}
\label{sec:experiments}

\subsection{Implementation Details}
\label{sec:experiments:details}

\paragraph{Datasets and Benchmarks.}
We adopt the popular DL3DV-10K dataset~\cite{ling2024dl3dv} as training set for Reliev3R. 
DL3DV-10K is a large-scale dataset with 10K scenes and 3M+ image frames, containing scenes of both indoor, outdoor and at different day time. 
To fully evaluate the performance of Reliev3R and its sibling models, we perform evaluation on DL3DV-benchmark dataset~\cite{ling2024dl3dv} for in-domain generalization performance and ScanNet++ dataset~\cite{yeshwanth2023scannet++} for out-of-domain zero-shot performance. 
Multi-view geometric annotations of DL3DV-benchmark dataset are produced with the open-sourced annotation pipeline of Map-Anything~\cite{keetha2025mapanything}. 

\vspace{-4mm}
\paragraph{Model Details.}
The architecture of Reliev3R is implemented based on $\pi^3$~\cite{wang2025pi} by replacing the point map prediction head with a depth prediction head. 
The Reliev3R model reported in this paper has 450M parameters in total. 
To produce pseudo label for monocular relative depth and image correspondence, Depth Pro~\cite{bochkovskii2024depth} and CoTracker~\cite{karaev2024cotracker} are introduced as pretrained pseudo labeling models. 
There are two points to notice. 
First, despite Depth Pro is capable to predict metric depth, we only use its prediction for relative depth regularization (see \cref{eq:4}). 
Second, neither Depth Pro and CoTracker are trained on DL3DV-10K dataset, which strictly simulates the real situation where we scale up the training dataset with data unseen by pseudo labeling models.

\vspace{-4mm}
\paragraph{Training Details.}
We train Reliev3R for 60k steps on 64 Ascend 910B3 NPUs for 3 days, with a total batch size of 64 and 8 image views given as input for each batch. 
The training starts with a learning rate of $10^{-4}$ which decays following a cosine scheduler. 

\subsection{Comparison with Prior FFRMs}
\label{sec:experiments:quantitative}

\paragraph{Baselines.}
To precisely evaluate the performance gap in reconstruction accuracy between Reliev3R and other FFRMs supervised with multi-view geometric annotations, we train $\pi^3$~\cite{wang2025pi} from scratch on a consistent training set with Reliev3R (\ie, using only DL3DV-10K dataset~\cite{ling2024dl3dv}), which is denoted as $\pi^{3\dag}$. 
We also introduce the official release of various FFRMs for evaluation. 
The latest SOTAs are colored \textcolor{lightgray}{gray} because Reliev3R is not comparable with them for training on only DL3DV-10K dataset, \textit{merely 5\%} of the data used by SOTAs. 
Meanwhile, we also compare the performance of Reliev3R with a feed-forward camera estimation model AnyCam~\cite{wimbauer2025anycam}, which is supervised without any multi-view geometric annotations as well. 
Inputting images, metric depth maps and optical flows, AnyCam predicts camera parameters which are supposed to register the input metric depth maps in the world coordinate the best.
This makes the comparison feasible between AnyCam and FFRMs. 

\vspace{-4mm}
\paragraph{Results.} 
We report the metrics on 8-view reconstruction for DL3DV-benchmark dataset in \cref{tab:main-result}, evaluating point map reconstruction, camera pose estimation and relative depth estimation. 
The input views are sampled with contiguous image frames. 
Although there is a distance for Reliev3R to catch up with FFRM SOTAs, Reliev3R is on par with and even outperforms early FFRMs such as MVDUSt3R~\cite{tang2025mv} and FLARE~\cite{zhang2025flare}. 
Furthermore, Reliev3R outperforms AnyCam significantly, demonstrating its superiority as a weakly-sepervised feed-forward 3D method. 
Please refer to \cref{fig:teaser} for qualitative comparison. 

\begin{table}[t]
    \centering
    \setlength\tabcolsep{4pt}
    \footnotesize
    \captionsetup{singlelinecheck=false}
    \caption{
        Zero-shot evaluation on ScanNet++~\cite{yeshwanth2023scannet++} dataset, which has a different focal length with DL3DV-10K dataset~\cite{ling2024dl3dv} that unseen during the training of Reliev3R and $\pi^{3\dag}$.
        The same metrics are reported as in \cref{tab:main-result}. 
        While consistently outperforming AnyCam~\cite{wimbauer2025anycam}, Reliev3R matches against $\pi^{3\dag}$ in overall performance and holds significant advantage in depth prediction. 
    }
    \begin{tabular}{c cc cc cc}
        \toprule
        {\multirow{2}{2cm}{\centering Method}} & \multicolumn{2}{c}{Point Map} & \multicolumn{2}{c}{Camera Pose} & \multicolumn{2}{c}{Depth Map} \\
                                                 & rel $\downarrow$ & $\tau\uparrow$ &  ATE $\downarrow$ & AUC $\uparrow$ & rel $\downarrow$ & $\tau\uparrow$ \\

        \midrule

        \multicolumn{1}{l}{\textcolor{lightgray}{$\pi^{3}$~\cite{wang2025pi}}}      &          \textcolor{lightgray}{0.027} &         \textcolor{lightgray}{0.972} &           \textcolor{lightgray}{0.002} &        \textcolor{lightgray}{86.941} &        \textcolor{lightgray}{0.027} &         \textcolor{lightgray}{0.970} \\
        \multicolumn{1}{l}{$\pi^{3\dag}$}                    &          \cellcolor{orange!50}0.232 &         \cellcolor{red!50}0.678 &           \cellcolor{red!50}0.028 &        \cellcolor{red!50}17.789 &        \cellcolor{orange!50}0.220 &         \cellcolor{orange!50}0.171 \\

        \hline
        
        \multicolumn{1}{l}{AnyCam~\cite{wimbauer2025anycam}} &          \cellcolor{yellow!20}0.438 &         \cellcolor{yellow!20}0.448 &           \cellcolor{yellow!20}0.084 &        \cellcolor{yellow!20}15.376 &        \cellcolor{yellow!20}0.286 &         \cellcolor{yellow!20}0.095 \\
        \multicolumn{1}{l}{Reliev3R(ours)}                   &          \cellcolor{red!50}0.172 &         \cellcolor{orange!50}0.594 &           \cellcolor{orange!50}0.030 &        \cellcolor{orange!50}15.711 &        \cellcolor{red!50}0.124 &         \cellcolor{red!50}0.583 \\
        
        \bottomrule
    \end{tabular}
    \label{tab:zero-shot}
\end{table}

\subsection{Zero-Shot Evaluation}
\label{sec:experiments:zeroshot}

To clearly depict the characteristics of Reliev3R compared with fully-supervised FFRMs, we perform evaluation on ScanNet dataset++~\cite{yeshwanth2023scannet++}, which has a different focal length and width-height ratio against DL3DV-10K dataset~\cite{ling2024dl3dv}. 
This can be proven by comparing the results of $\pi^{3\dag}$ and $\pi^{3}$~\cite{wang2025pi}, where ScanNet++ dataset is contained within the training set of $\pi^3$. 
As one of the baselines, we introduce $\pi^{3\dag}$ referred in the ~\cref{sec:experiments:quantitative}, which shares the same training configuration with Reliev3R but fully-supervised with multi-view geometric annotations. 
We also introduce AnyCam~\cite{wimbauer2025anycam} and official $\pi^3$~\cite{wang2025pi} as baselines for weakly-supervised feed-forward 3D model and SOTA fully-supervised SSFM, respectively. 
As shown in \cref{tab:zero-shot}, both Reliev3R and $\pi^{3\dag}$ overfits on the focal length of DL3DV-10K dataset and drop significantly in reconstruction precision on ScanNet++ dataset. 
Despite of the overfitting issue, Reliev3R holds its advantage in reconstruction precision over AnyCam. 
What's worth noticing is that, Reliev3R is on par with $\pi^{3\dag}$ in this zero-shot evaluation experiment, even significantly surpassing $\pi^{3\dag}$ in depth estimation. 
This experiment demonstrates Reliev3R to be more robust than fully-supervised FFRMs against zero-shot prediction.

\begin{figure*}[t]
    \centering
    \captionsetup{singlelinecheck=false}
    \includegraphics[width=\linewidth]{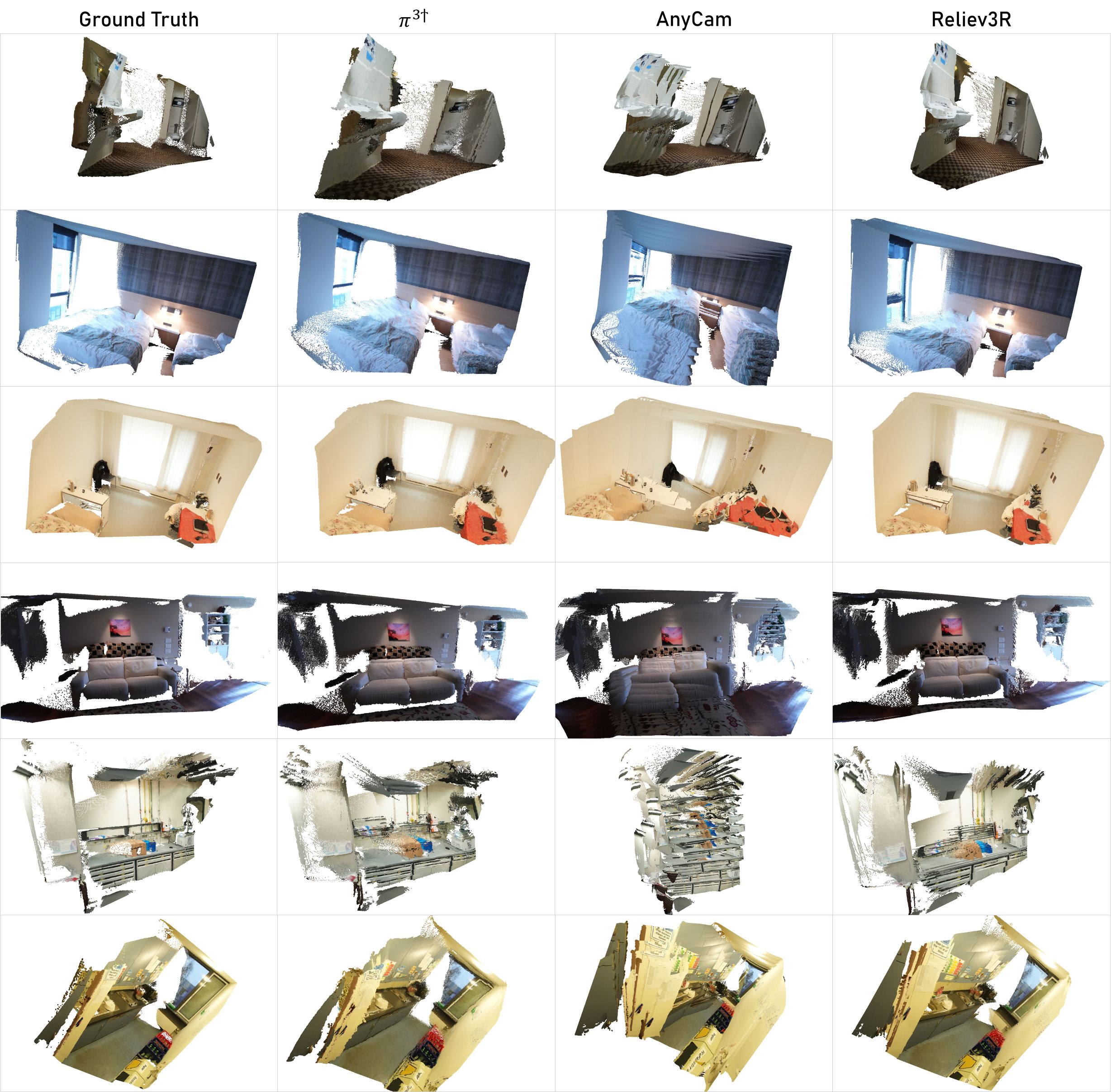}
    \caption{
        Visualization of point maps for zero-shot evaluation on ScanNet++ dataset~\cite{dai2017scannet}, which has a different focal length from DL3DV-10K dataest~\cite{ling2024dl3dv}.
        Reliev3R holds its advantage over AnyCam~\cite{wimbauer2025anycam} even in zero-shot circumstance. 
        Furthermore, Reliev3R presents a notable zero-shot performance on par with fully-supervised $\pi^{3\dag}$ training on the same datasets.
    }
    \label{fig:zeroshot}
\end{figure*}

\section{Limitation}
\label{sec:limitaion}

Despite of the superiority and promising potential of Reliev3R, this study presents several limitations that highlight directions for future research.
The major limitation lies in the absence of a large-scale data scaling analysis. 
Although Reliev3R is designed to reduce the reliance on SfM/MVS annotations and thereby increase the feasible training scale, we have not empirically verified its behavior under substantially larger datasets.
Another limitation concerns dynamic scenes. 
Our current formulation does not explicitly handle multi-view inconsistencies cause by dynamics, which may limit its applicability to unconstrained, in-the-wild video collections—an important setting for truly unlimited data scaling. 
Finally, our approach inherits certain weaknesses from the pseudo-label sources we rely upon. 
These aspects are crucial for ensuring robustness when scaling training to more heterogeneous datasets.
\section{Conclusion}
\label{sec:conclusion}

In this paper, we present Reliev3R, the first weakly-supervised paradigm that enables training feed-forward reconstruction models entirely from scratch without any multi-view geometric annotations. 
By designing a dedicated training objective based on multi-view geometric constraints, Reliev3R succeeds to learn predicting depth maps and camera poses that are registered in 3D world coordinate. 
Extensive comparison with baselines demonstrate that Reliev3R surpasses early fully-supervised FFRMs and is comparable with the SOTAs, holding the potential to fully release FFRMs from multi-view geometric annotations with data scaling up. 
Future improvements of Reliev3R lies in scaling up of training data and adapting to reconstruction from dynamic image groups. 
We hope this study may draw the attention of the community to investigate a supervision paradigm which is easily scalable for further development of FFRMs. 

{
    \small
    \bibliographystyle{ieeenat_fullname}
    \bibliography{main}
}
\clearpage
\maketitlesupplementary

\appendix

\section{Training Setup for Baselines}
\label{sec:supp:exp_detail}

We introduce the training setup of baseline methods in this section for better depicting advantages of Reliev3R. 

\vspace{-4mm}
\paragraph{CUT3R, VGGT, $\pi^3$, and MapAnything~\cite{wang2025continuous, wang2025vggt, wang2025pi, keetha2025mapanything}}.
These works are representatives as the edge of FFRMs. 
Each of these FFRM is trained on 15-17 diverse datasets with multi-view geometric annotations, which include approximately 200k scenes in total.

\vspace{-4mm}
\paragraph{$\mathbf{\pi^{3\dag}}$~\cite{wang2025pi}.}
For a comprehensive comparison between Reliev3R and SOTA FFRMs, we train $\pi^3$~\cite{wang2025pi} from scratch with the same training set as Reliev3R, \ie the DL3DV-10K~\cite{ling2024dl3dv} dataset. 
The result is denoted as $\pi^{3\dag}$. 
\textbf{\textit{The parameter size of $\pi^{3\dag}$ is controlled approximate to Reliev3R as 450M. }}
The training for $\pi^{3\dag}$ is directed with the same learning rate schedule as Reliev3R, and optimized for 30k steps until a full convergence on the validation set, which is randomly sampled as 2\% of the DL3DV-10K dataset. 
Other setups, such as training objectives, follow $\pi^3$~\cite{wang2025pi} by default.

\vspace{-4mm}
\paragraph{MVDUSt3R~\cite{tang2025mv}.}
MVDUSt3R is one of the pioneering works to expand the input of DUSt3R~\cite{wang2024dust3r} from two views to multiple views. 
MVDUSt3R is trained by finetuning DUSt3R on the subset of its training data, which is relatively small compared to the training data of SOTA FFRMs. 
As a result, MVDUSt3R is hindered with overfitting issue when performing a zero-shot evaluation on unseen datasets such as DL3DV-benchmark~\cite{ling2024dl3dv}, which explains the catastrophic performance of MVDUSt3R in the paper. 

\vspace{-4mm}
\paragraph{FLARE~\cite{zhang2025flare}.}
Similar to MVDUst3R~\cite{tang2025mv}, FLARE is one of the early works to expand the input of DUSt3R~\cite{wang2024dust3r} to multiple views. 
Differently, FLARE is capable to directly predict the camera pose along the forward pass of the model. 
\textit{\textbf{We adopt the direct output from FLARE model as its camera pose prediction, instead of following DUSt3R~\cite{wang2024dust3r} to post process the predicted point maps with a PnP solver~\cite{fischler1981random} to estimate camera poses. }}
As shown in the paper (both figures and tables), we find that the directly predicted camera pose of FLARE has an inconsistent scale with the point maps.
This is because, to handle the unreliability in reprojection when camera pose is learned from scratch, FLARE proposes a learnable reprojector conditioned on the learned camera poses, without further constraint on the scale consistency between the reprojected point maps and camera positions. 
The unreliability of reprojection is addressed by Reliev3R with a trigonometry-based reprojection loss, which is capable to produce meaningful gradient for optimization given any random camera poses. 

\vspace{-4mm}
\paragraph{AnyCam~\cite{wimbauer2025anycam}.}
Since there is not a proper baseline to be compared with Reliev3R as a weakly-supervised FFRM, we introduce AnyCam as a less rigorous baseline. 
AnyCam is one of the few works to explore training a feed-forward 3D model without multi-view geometric annotations. 
More concretely, AnyCam focuses on estimating both intrinsics and poses of cameras given a group of input views followed by view-wise metric depth maps and optical flows. 
By back-projecting the input metric depth maps into the world coordinate with predicted camera parameters, a registered point cloud can be produced to be compared with FFRMs. 
For a fair comparison, we condition the camera pose prediction of AnyCam on the ground truth focal length of cameras. 

\begin{table}[t]
    \centering
    \setlength\tabcolsep{4pt}
    \footnotesize
    \captionsetup{singlelinecheck=false}
    \caption{
        We train $\pi^3$ from-scratch (random initial weights) in a fully-supervised manner on a down-sampled mixture of synthetic datasets, including ASE dataset, TartanAir dataset, MVS-Synth dataset, Blended-MVS dataset and \etc following VGGT. 
        The training runs for 20k iterations with a total batch size of 128.
        The result is denoted as `Synth Fully-Sup'. in the table. 
        Then we subsequently finetune this checkpoint on DL3DV dataset for 20k steps using the weak supervision proposed by Reliev3R. 
        The result is denoted as `+Reliev3R' in the table. 
        We report the same metrics as Table 1 in the paper, which are evaluated on DL3DV-benchmark dataset. 
        We also evaluate camera pose estimation for RayZer on this benchmark with its recently open-sourced checkpoint. 
    }
    \begin{tabular}{c cc cc cc}
        \toprule
        {\multirow{2}{2cm}{\centering Method}} & \multicolumn{2}{c}{Point Map} & \multicolumn{2}{c}{Camera Pose} & \multicolumn{2}{c}{Depth Map} \\
                                                 & rel $\downarrow$ & $\tau\uparrow$ &  ATE $\downarrow$ & AUC $\uparrow$ & rel $\downarrow$ & $\tau\uparrow$ \\

        \midrule
        \multicolumn{1}{l}{RayZer}                &           $\times$ &          $\times$ &           0.786 &         0.362 &         $\times$ &          $\times$ \\
        \midrule
        \multicolumn{1}{l}{Synth Fully-Sup.}      &          0.277 &         0.475 &           0.042 &        17.075 &        0.208 &         0.357 \\
        \multicolumn{1}{l}{+Reliev3R}             &          \textbf{0.137} &         \textbf{0.667} &           \textbf{0.033} &        \textbf{34.240} &        
        \textbf{0.106} &         \textbf{0.679} \\
        
        \bottomrule
    \end{tabular}
    \label{tab:rebu}
\end{table}

\section{Extending Reliev3R to Semi-Supervision}
\label{sec:supp:semi}

It would be a practical approach to utilize Reliev3R to generalize a FFRM, which is fully-supervised on synthetic data, to realistic data, where ground-truth is hard to access. 
As a proof of concept, we perform an experiment in \cref{tab:rebu}. 
The FFRM trained merely on synthetic data performs bad on DL3DV because of domain gap between realistic data and synthetic data. 
Reliev3R succeeds to improve this FFRM, demonstrating the feasibility of the warm-up\&finetune usage of Reliev3R. 
In \cref{tab:rebu}, both `Synth Fully-Sup.' and `+Reliev3R' are equipped with a FoV head for intrinsic prediction, where the former is supervised with ground-truth intrinsics and the latter is supervised with pseudo FoV predicted by MoGe2. 
To avoid overfitting on FoV, we introduce image cropping as augmentation to FoV. 
The pseudo depth labels used to perform the experiment in \cref{tab:rebu} are generated by MoGe2 to keep consistency with pseudo intrinsics.

\begin{figure*}[h!]
    \centering
    \captionsetup{singlelinecheck=false}
    \vspace{30mm}
    \includegraphics[width=\linewidth]{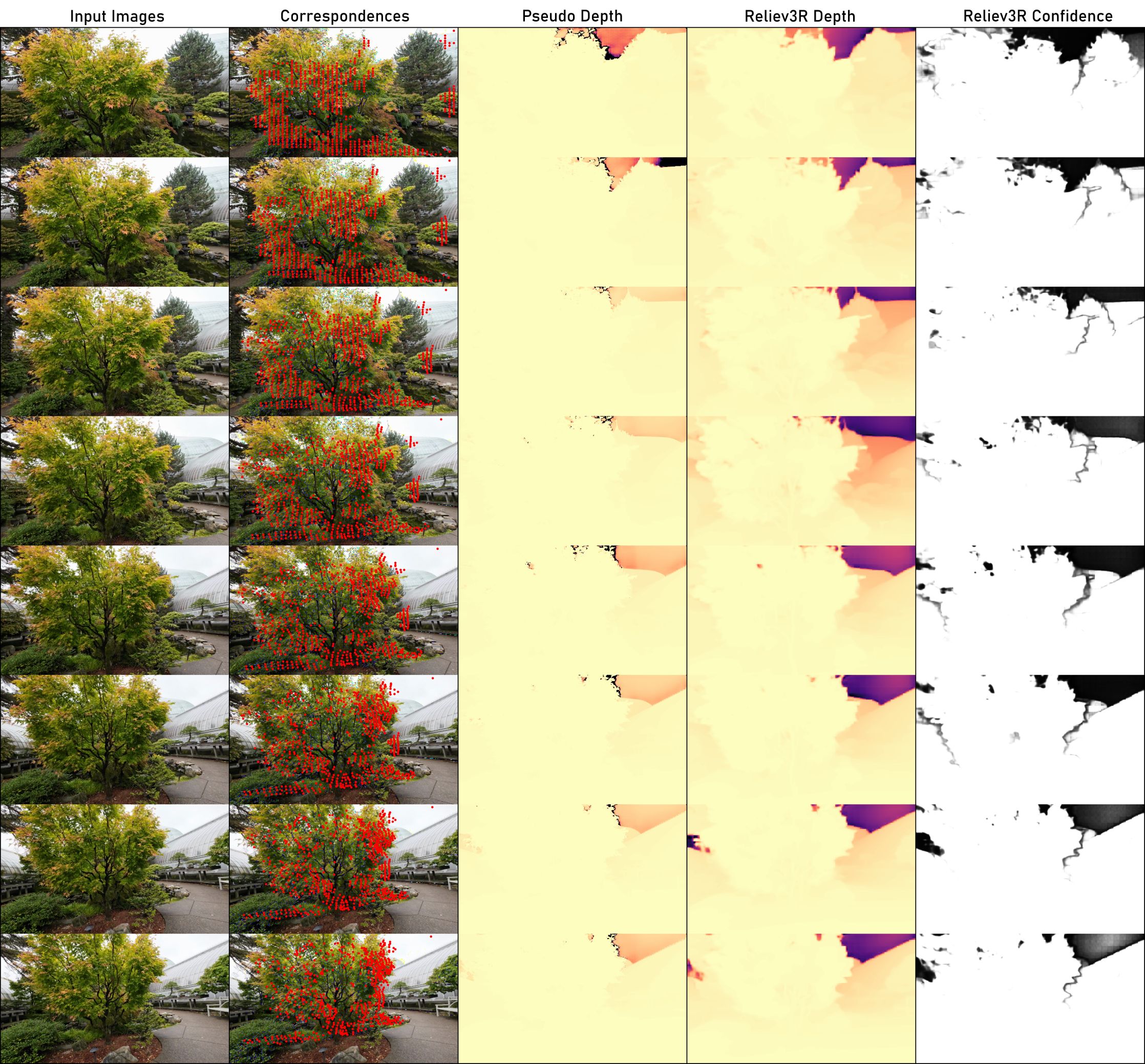}
    \caption{
        Visualization of input (multi-view images), output (multi-view depth maps and confidence maps) and pseudo labels (multi-view correspondences, monocular depth maps) of Reliev3R. 
    }
    \vspace{20mm}
    \label{fig:supp:figure3-1}
\end{figure*}

\begin{figure*}[h!]
    \centering
    \captionsetup{singlelinecheck=false}
    \vspace{30mm}
    \includegraphics[width=\linewidth]{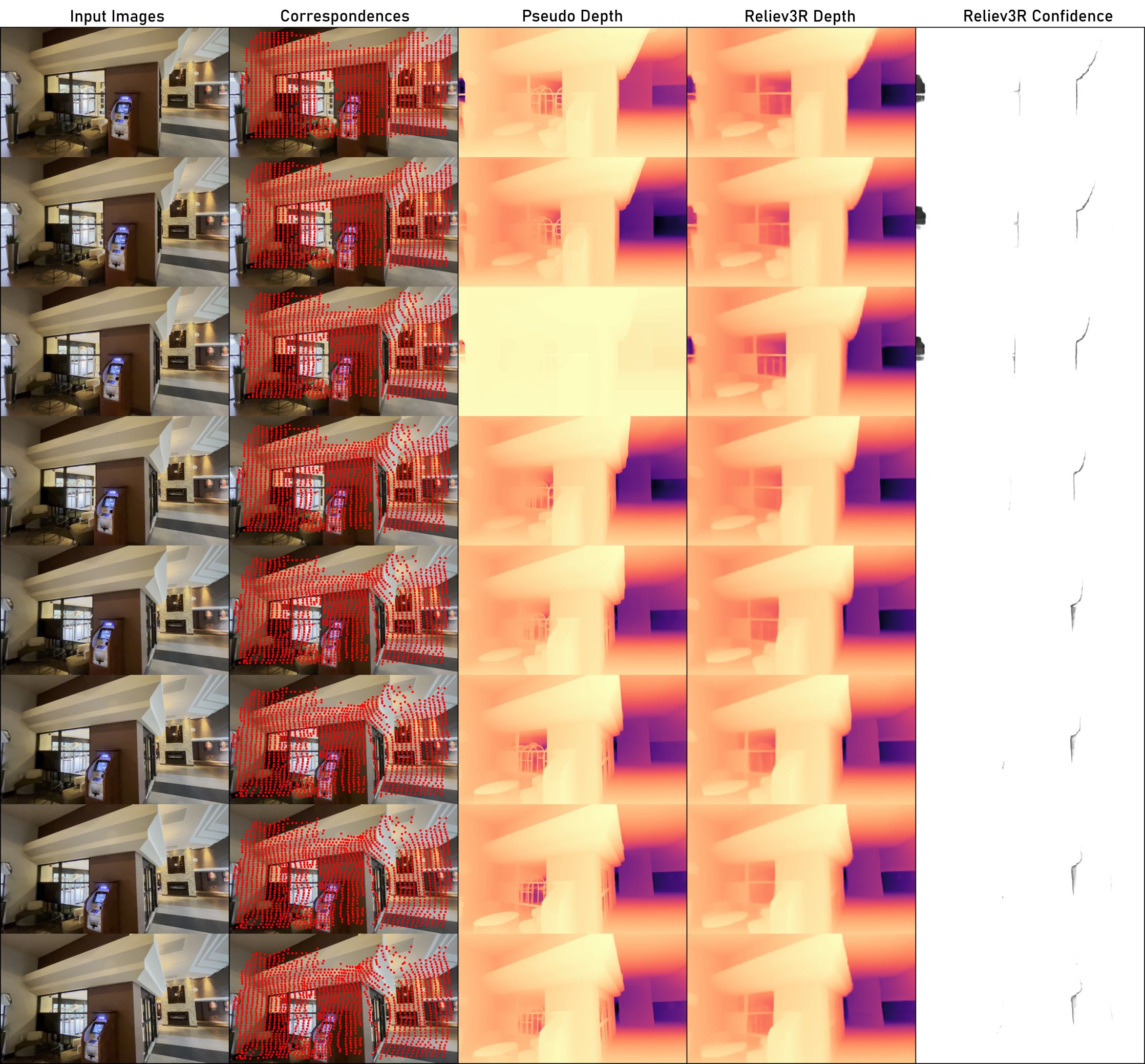}
    \caption{
        Visualization of input (multi-view images), output (multi-view depth maps and confidence maps) and pseudo labels (multi-view correspondences, monocular depth maps) of Reliev3R. 
    }
    \vspace{20mm}
    \label{fig:supp:figure3-2}
\end{figure*}

\begin{figure*}[h!]
    \centering
    \captionsetup{singlelinecheck=false}
    \vspace{30mm}
    \includegraphics[width=\linewidth]{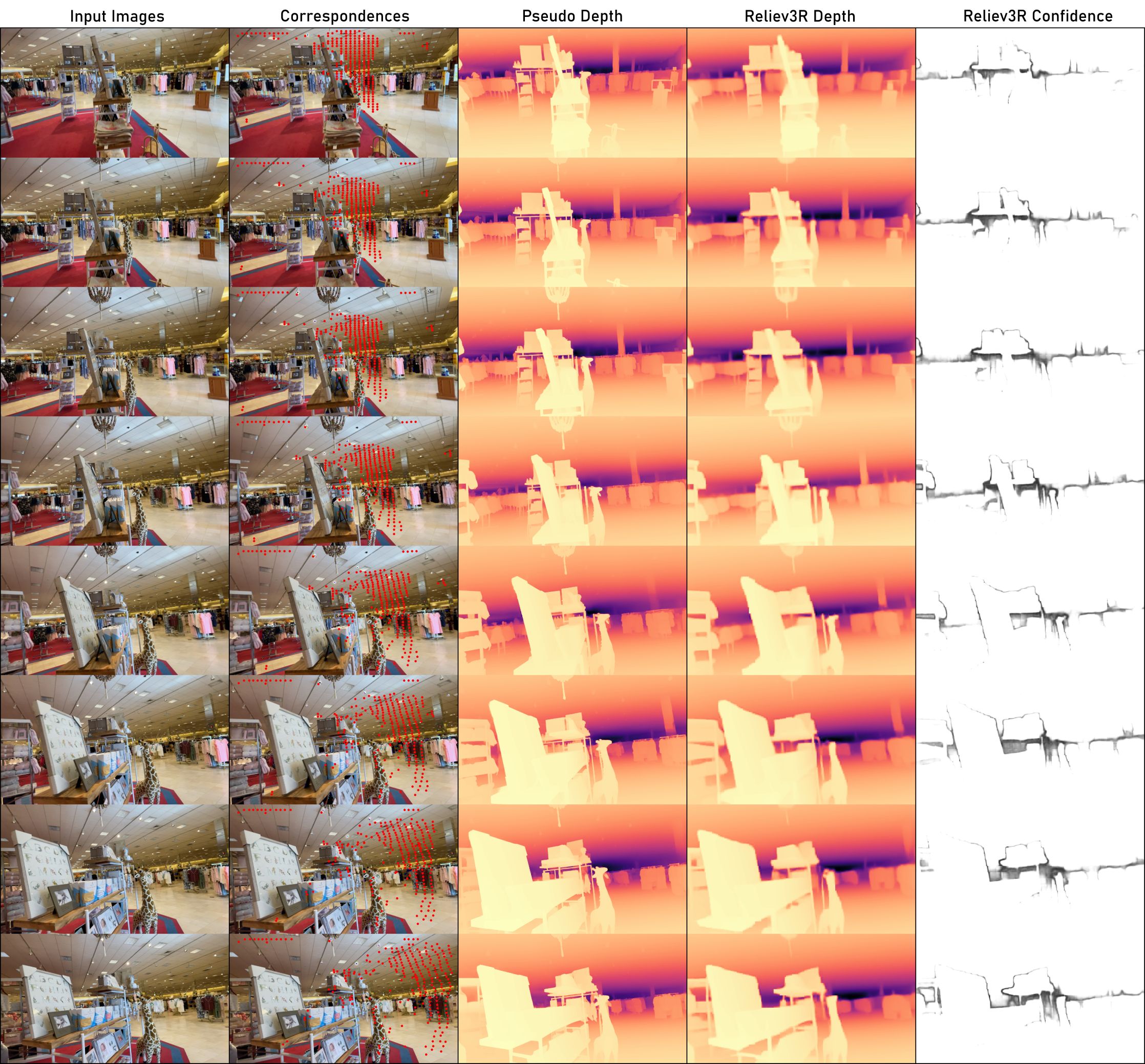}
    \caption{
        Visualization of input (multi-view images), output (multi-view depth maps and confidence maps) and pseudo labels (multi-view correspondences, monocular depth maps) of Reliev3R. 
    }
    \vspace{20mm}
    \label{fig:supp:figure3-3}
\end{figure*}

\begin{figure*}[h!]
    \centering
    \captionsetup{singlelinecheck=false}
    \vspace{30mm}
    \includegraphics[width=\linewidth]{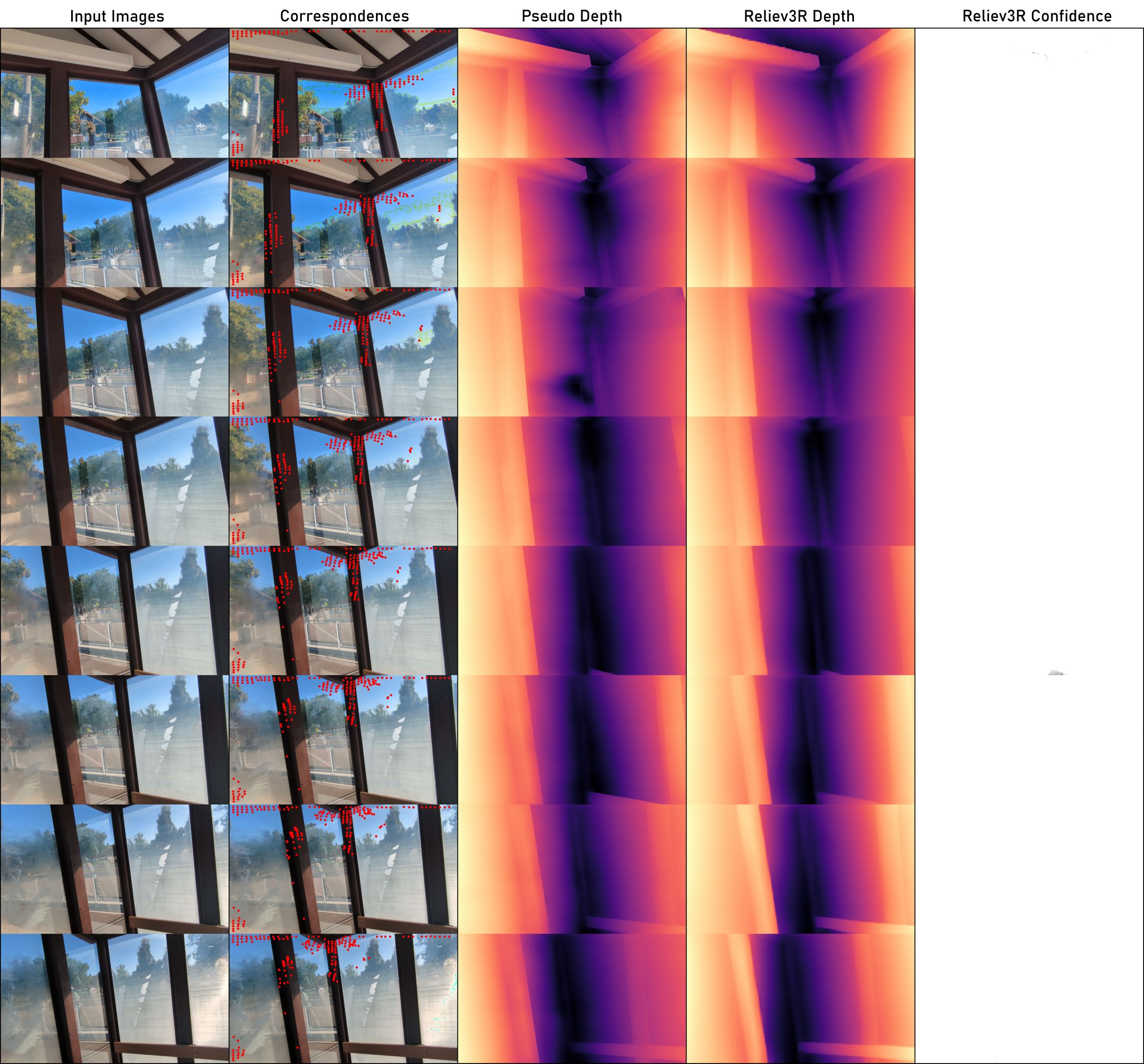}
    \caption{
        Visualization of input (multi-view images), output (multi-view depth maps and confidence maps) and pseudo labels (multi-view correspondences, monocular depth maps) of Reliev3R. 
    }
    \vspace{20mm}
    \label{fig:supp:figure3-4}
\end{figure*}

\begin{table}[t]
    \centering
    \setlength\tabcolsep{4pt}
    \footnotesize
    \captionsetup{singlelinecheck=false}
    \caption{
        Ablation study on different confidence weight $\alpha$ on DL3DV-benchmark~\cite{ling2024dl3dv} dataset. 
        $\alpha=1.0$ is adopted in the paper to train Reliev3R. 
        Absolute relative error (rel) and inlier ratio at a relative threshold of 10\% ($\tau$) are reported to evaluate point maps and depth maps. 
        Average aligned trajectory error (ATE) and area under curve at an error threshold of 30\degree{} (AUC) are reported to evluate camera pose estimation. 
    }
    \begin{tabular}{c cc cc cc}
        \toprule
        {\multirow{2}{2cm}{\centering }} & \multicolumn{2}{c}{Point Map} & \multicolumn{2}{c}{Camera Pose} & \multicolumn{2}{c}{Depth Map} \\
                                                 & rel $\downarrow$ & $\tau\uparrow$ &  ATE $\downarrow$ & AUC $\uparrow$ & rel $\downarrow$ & $\tau\uparrow$ \\

        \midrule

        \multicolumn{1}{l}{$\alpha=0.2$}            &          0.143 &          0.607 &          0.022 &          42.116 &          0.137 &          0.606 \\
        \multicolumn{1}{l}{$\alpha=0.5$}            &          0.137 &          0.626 &          0.023 &          46.703 &          0.129 &          0.628 \\
        \multicolumn{1}{l}{$\alpha=1.0$ (Reliev3R)} & \textbf{0.122} & \textbf{0.663} & \textbf{0.018} & \textbf{49.426} & \textbf{0.115} & \textbf{0.657} \\
        \multicolumn{1}{l}{$\alpha=2.0$}            &          0.127 &          0.651 &          0.020 &          48.566 &          0.121 &          0.647 \\
        
        \bottomrule
    \end{tabular}
    \label{tab:supp:ablation}
\end{table}

\section{Ablation Study}
\label{sec:supp:ablation}

In Eq.4 of the paper, we introduce an ambiguity-aware scale-invariant depth loss to regularize the shape of Reliev3R depth prediction while addressing the multi-view inconsistency in pseudo monocular depth labels. 
Intuitively, a large $\alpha$ encourages the training of Reliev3R to ignore multi-view inconsistency, and ends up with the confidence learned as a constant of 2 (the confidence value is restricted within $(0, 2)$).
On the other hand, a small $\alpha$ results in degraded camera pose estimation and depth registration for disabling the depth shape regularization. 

To evaluate the sensitivity of Reliev3R to $\alpha$, we present ablation study on different $\alpha$, as shown in \cref{tab:supp:ablation}. 
The results advocate the analysis above, and demonstrate that Reliev3R is robust against a large variation in $\alpha$.

\section{More Visualization for Reliev3R Prediction}
\label{sec:supp:more_vis}
We show the input, output and pseudo labels of Reliev3R in \cref{fig:supp:figure3-1}, \cref{fig:supp:figure3-2}, \cref{fig:supp:figure3-3} and \cref{fig:supp:figure3-4}. 
It's observed that Reliev3R surpasses the pseudo monocular depth labels in case of multi-view depth consistency. 
And the learned confidence maps function well as we designed to mask out unreliable regions such as sky, contents in the far and edges of objects.


\end{document}